\pdfoutput=1

\documentclass[11pt]{article}

\usepackage{acl}

\usepackage{times}
\usepackage{geometry}
\usepackage{multirow, makecell}
\usepackage{rotating}
\usepackage{siunitx}
\usepackage{latexsym}
\usepackage{mathtools}
\usepackage{amssymb}
\usepackage{amsfonts}
\usepackage{caption}
\usepackage[T1]{fontenc}

\usepackage[utf8]{inputenc}
\usepackage{amsmath}
\usepackage{graphicx}
\usepackage{microtype}

\usepackage{CJKutf8}

\title{Document-Level Relation Extraction \\ with Adaptive Focal Loss and Knowledge Distillation}

\author{Qingyu Tan\thanks{$^\dag$  Qingyu Tan is under the Joint PhD Program between Alibaba and National University of Singapore.}$^{~1,2}$ \quad \textbf{Ruidan He\thanks{$^\dag$  Corresponding author}$^{~1}$ \quad Lidong Bing$^{1}$ \quad Hwee Tou Ng$^{2}$} 
\\$^1$DAMO Academy, Alibaba Group~~\\
$^2$Department of Computer Science, National University of Singapore\\
\texttt{\{qingyu.tan,ruidan.he,l.bing\}@alibaba-inc.com}\\
\texttt{\{qtan6,nght\}@comp.nus.edu.sg}\\
}

\begin{document}
\maketitle
\begin{abstract}
Document-level Relation Extraction (DocRE) is a more challenging task compared to its sentence-level counterpart. It aims to extract relations from multiple sentences at once. In this paper, we propose a semi-supervised framework for DocRE with three novel components. Firstly, we use an axial attention module for learning the interdependency among entity-pairs, which improves the performance on two-hop relations. Secondly, we propose an adaptive focal loss to tackle the class imbalance problem of DocRE. Lastly, we use knowledge distillation to overcome the differences between human annotated data and distantly supervised data. We conducted experiments on two DocRE datasets. Our model consistently outperforms strong baselines and its performance exceeds the previous SOTA by 1.36 F1 and 1.46 Ign\_F1 score on the DocRED leaderboard.\footnote{Our code and data are available at {\url{https://github.com/tonytan48/KD-DocRE}}} 
\end{abstract}


\section{Introduction}


The problem of document-level relation extraction\footnote{In this work, the task of relation extraction presumes that entities are given.} (DocRE) is highly important for information extraction and NLP research. The DocRE task aims to extract relations among multiple entities within a document. The DocRE task is more challenging than its sentence-level counterpart in the following aspects: (1) The complexity of DocRE increases quadratically with the number of entities. If a document contains $n$ entities, classification decisions must be made on $n(n-1)$ entity pairs and most of them do not contain any relation. (2) Aside from the imbalance of positive and negative examples, the distribution of relation types for the positive entity pairs is also highly imbalanced. Considering the DocRED \citep{yao2019docred} dataset as an example, there are 96 relation types in total, where the top 10 relations take up 59.4\% of all the relation labels. This imbalance significantly increases the difficulty of the document-level RE task.




Most existing approaches of DocRE leverage dependency information to construct a document-level graph (\citealp{zeng2021sire}; \citealp{zeng2020double}), and then use graph neural networks for reasoning. Another popular strand of this field uses transformer-only~\citep{vaswani2017attention} architecture~(\citealp{zhou2021document}; \citealp{xu2021entity}; \citealp{zhang2021document}). Such models are able to achieve state-of-the-art performance without explicit graph reasoning, showing that pre-trained language models (PrLMs) are able to implicitly capture long-distance relationships.  However, there are three limitations of the existing DocRE methods. Firstly, existing methods mainly focus on the syntactic features from PrLMs while neglecting the interactions between entity pairs. \citet{zhang2021document} and \citet{li2021mrn} have used CNN structure to encode the interaction between entity pairs, but CNN structure cannot capture all the elements within the two-hop reasoning paths. Secondly, there is no prior work that explicitly tackles the class-imbalance problem for DocRE. Existing works~(\citealp{zhou2021document}; \citealp{zhang2021document}; \citealp{zeng2020double}) only focus on threshold learning for balancing the positive and negative examples, but the class-imbalance problem within positive examples is not addressed. Lastly, there are very few works discussing the method of adapting distantly supervised data for the DocRE task. \citet{xu2021entity} has shown that distantly supervised data is able to improve the performance of document-level relation extraction. However, it only uses the distantly supervised data to pre-train the RE model in a naive manner.
\begin{figure*}
    \centering
    \resizebox{0.85\textwidth}{!}{
    \includegraphics{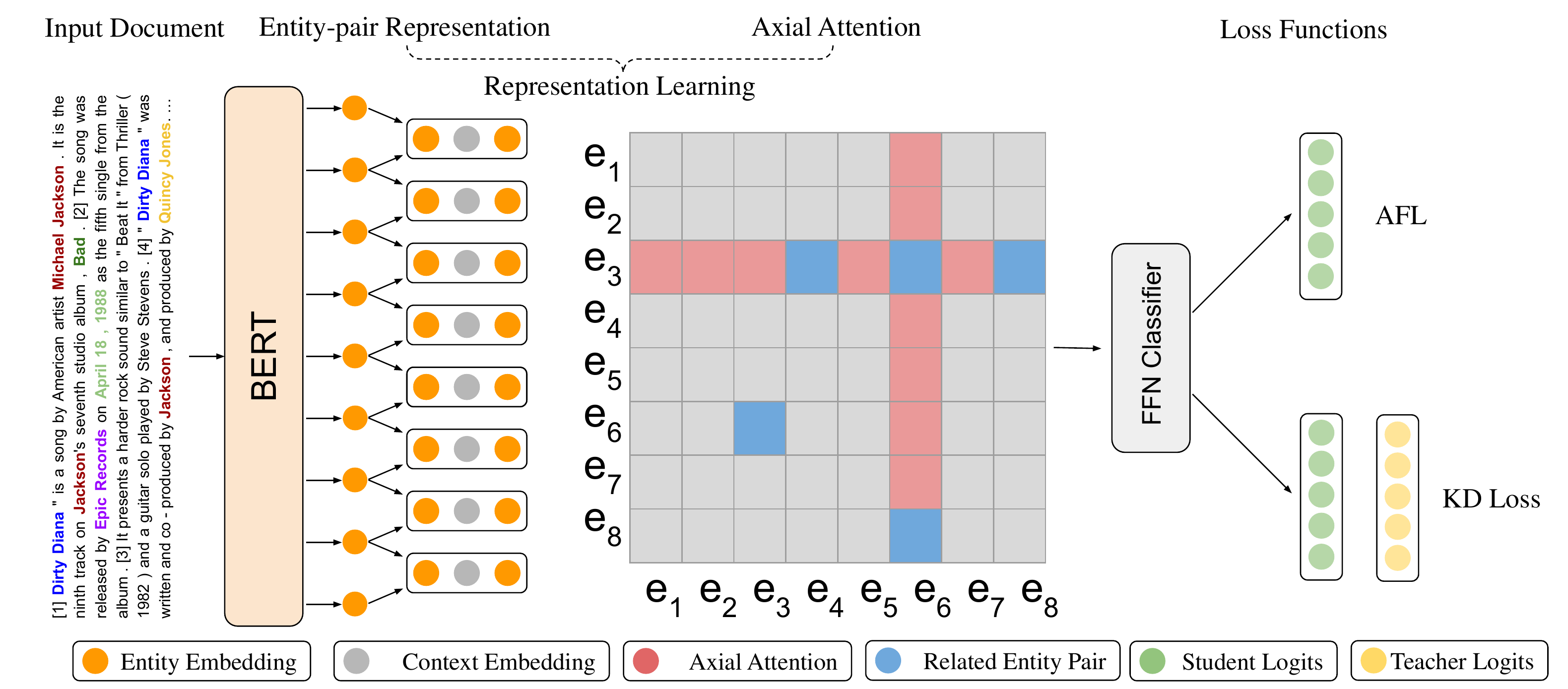}
    }
    \setlength{\abovecaptionskip}{0pt} 
    \captionsetup{margin=2.5em}
    \caption{Model architecture of our DocRE system. We show the axial attention region for the entity pair $(e_3, e_6)$.}
    \label{fig:model-architecture}
    
\end{figure*}
%

To overcome the limitations of existing works, we propose a semi-supervised learning framework for document-level relation extraction. Firstly, to improve the reasoning for two-hop relations, we propose to use an axial attention module as feature extractor. This module enables us to attend to elements that are within two-hop logical paths and capture the interdependency among the relation triplets. Secondly, we propose Adaptive Focal Loss to address the imbalanced label distribution problem. The proposed loss function encourages the long-tail classes to contribute more to the overall loss. Lastly, we use knowledge distillation to overcome the differences between the annotated data and the distantly supervised data. Specifically, we first train a teacher model with a small amount of human annotated data. The teacher model will then be used to generate predictions on a large amount of distantly supervised data. The generated predictions are used as soft labels for pre-training our student model. Finally, the pre-trained student model is further fine-tuned on the human annotated data.

We conducted experiments on two datasets -- the DocRED \citep{yao2019docred} dataset and the HacRED \citep{cheng-etal-2021-hacred} dataset. Experimental results show that our model consistently outperforms competitive baselines. Moreover, our model significantly outperforms the existing state-of-the-art \textbf{SSAN-Adapt} \citep{xu2021entity} on the DocRED leaderboard by 1.36 in F1 score and 1.46 in Ign\_F1 score.\footnote{Refer to {\url{https://competitions.codalab.org/competitions/20717}}, where our model is named \textit{KD-Roberta}.} Besides, we provide a thorough ablation study and error analysis to identify the bottleneck of our method. 




\section{Methodology}
\subsection{Problem Formulation}
In this section, we describe the task formulation of document-level relation classification. Given a document $D$ that contains a set of entities $\{e_i\}_{i=1}^n$, the document-level relation extraction task is to predict the relation types between entity pairs $(e_s,e_o)_{s,o \in \{1...n\}, {s}\neq{o}}$, where the subscripts of $e_s$ and $e_o$ refer to subject and object. The set of relations is defined as $\textbf{R} \cup \{\textbf{NR}\}$, where \textbf{NR} stands for \textit{no relation}. An entity may occur multiple times in a document, thus for each entity $e_i$, there can be multiple mentions $\{m_j^{i}\}^{N_{e_i}}_{j=1}$. If no relation exists between the entities in the pair $(e_s,e_o)$, it will be labeled as \textbf{NR}. During test time, the relation labels for all entity pairs $(e_s,e_o)_{s,o \in \{1...n\}, {s}\neq{o}}$ will be predicted. Essentially, this is a multi-label classification problem, as there can be multiple relations between $e_s$ and $e_o$.




\subsection{Model Architecture}

As shown in Figure~\ref{fig:model-architecture}, our semi-supervised learning framework mainly consists of three parts: (1) representation learning; (2) adaptive focal loss; and (3) knowledge distillation for distant supervision pretraining. For representation learning, we first extract the contextual representation for each entity-pair by a pre-trained language model. The entity pair representations will be further enhanced by the axial attention module, which will encode the inter-dependent information between entity pairs. We then use a feedforward neural network (FFN) classifier to obtain the logits and compute their losses. We use our proposed adaptive focal loss to better learn from long-tail classes. Finally, we use knowledge distillation to overcome the differences between human annotated data and distantly supervised data. Specifically, we train a teacher model with the annotated data and use its output as soft labels. We then pre-train a student model based on the soft labels and the distant labels. The pre-trained student model will be fine-tuned again with the annotated data. We will describe the details for each part in the following sections.

\subsubsection{Representation Learning}
\paragraph{Entity Representation} We use a pretrained language model as the encoder. For a document $D$ of length $l$, we have $D = [x_{t}]_{t=1}^{l}$, where $x_{t}$ is the word at location $t$. Following prior works for relation classification, we use special token markers to represent entities. The entity mentions will be marked by a special token "*" at the start and end position. We then use a pre-trained language model (PrLM) to obtain the contextualized embeddings $\mathbf{H}$ of this document. 
\begin{equation}
    \mathbf{H} = PrLM([x_1,...,x_l]) = [h_1,...,h_l])
\end{equation}
where $\mathbf{H} \in \mathbb{R}^{{l}\times{d}}$ and $d$ is the hidden dimension of the PrLM. If the document length exceeds the maximum position of the PrLM, the document will be encoded as multiple overlapping chunks, and the contextualized embeddings of the overlapping chunks will be averaged. We take the embedding of the special token "*" at the start of the mention as its embedding, which is denoted as $h_{m_{j}}$. Then, for each entity $e_{i}$ with mentions $\{m_j^{i}\}^{N_{e_i}}_{j=1}$, where $N_{e_{i}}$ is the number of mentions for entity $e_{i}$, its global representation is obtained by logsumexp pooling:
\begin{equation}
    h_{e_{i}} = log\sum^{N_{e_{i}}}_{j=1}exp(h_{m_{j}})
\end{equation}
where $h_{e_{i}} \in \mathbb{R}^{d}$ is the aggregated feature of $e_{i}$.

\paragraph{Context-enhanced Entity Representation} As prior works (\citealp{xu2021entity}; \citealp{peng2020learning}) have shown that contextual information is crucial for the relation classification task, our model also adapts contextual pooling method from \citet{zhou2021document}. For each entity $e_{i}$, we first aggregate the attention output for its mentions by mean pooling $A_{e_{i}} = \sum^{N_{e_{i}}}_{j=1}(a_{m_{j}})$, where $a_{m_{j}} \in \mathbb{R}^{{H}\times{l}}$ is the the self-attention weight at the position of mention $m_{j}$, $H$ is the number of attention heads, and $l$ is the document length. Then the context query is calculated as:
\begin{equation}
    \abovedisplayskip=2pt
    \belowdisplayskip=2pt
    q^{(s,o)} = \sum_{i=1}^{H} ({A^{i}_{e_{s}}} \circ {A^{i}_{e_{o}}})
\end{equation}
\begin{equation}
    \abovedisplayskip=2pt
    \belowdisplayskip=2pt
    c^{(s,o)} = \mathbf{H}^{\intercal}q^{(s,o)}
\end{equation}
where $A_{e_{s}} \in \mathbb{R}^{{H}\times{l}}$ is the aggregated attention output for entity $e_{s}$, likewise for $e_{o}$. $q^{(s,o)} \in \mathbb{R}^{{l}}$ is the mean-pooled attention weight for entity pair $(e_{s}, e_{o})$ and $\mathbf{H} \in \mathbb{R}^{{l} \times {d}}$ is the contextual embedding of the whole document. Then the context vector $c^{(s,o)} \in \mathbb{R}^{d}$ is fused with the entity representations.
\begin{equation}
    z_{s} = \mathit{tanh}(\mathbf{W}_{s}h_{e_{s}} + \mathbf{W}_{c}c^{(s,o)})
\end{equation}
where $z_{s} \in \mathbb{R}^{d}$ is the context-enhanced representation of subject $s$ for entity pair $(e_s,e_o)$. We obtain the object representation $z_{o}$ in the same manner.

\paragraph{Entity Pair Representation} Following \citet{zhou2021document}, we use a grouped bilinear function for feature combination. The entity embedding $z_{s}$ will first will be split into $k$ equal-sized groups, such that $z_{s} = [z_{s}^{1}, z_{s}^{2}, ..., z_{s}^{k}]$. We perform the same splitting for $z_{o}$. The value $g^{(s,o)}_{i}$ at each dimension of our entity pair representation is obtained by:
\begin{equation}
\begin{split}
    g^{(s,o)}_{i} = \sum^{k}_{j=1}(z^{j\intercal}_{s}W^{j}_{g_{i}}z^{j}_{o}) +b_{i} \\
    g^{(s,o)}=[g^{(s,o)}_{1}, g^{(s,o)}_{2}, ..., g^{(s,o)}_{d}]
\end{split}
\end{equation}
where $W^{j}_{g_{i}} \in \mathbb{R}^{d/k \times d/k}$, for $i=1,...,d$, $j=1,...,k$, is the weight matrix for dimension $i$. $b_{i}$ is a scalar bias of dimension $i$. $g^{(s,o)} \in \mathbb{R}^{d}$ is our final entity pair representation. 

For a given document $D$ with $n$ entities, we need to classify $n(n-1)$ number of entity pair permutations. To help us encode all the entity pairs and their positions, we used an $\mathbb{R}^{{n}\times{n}\times{d}}$ matrix $\mathbf{G}$ to represent all the entity pairs of document $D$, and the diagonal of the ${n}\times{n}$ index is neglected during training and inference.


\paragraph{Axial Attention-Enhanced Entity Pair Representation} Instead of using only head and tail embedding for relation classification, we propose to use two-hop attention to encode the axial neighboring information of each entity pair $(e_s, e_o)$ representation. Although there are prior works that use Convolution Neural Networks (CNNs) to encode the neighbor information for relation classification (\citealp{zhang2021document}), we believe that attending to the axial elements is more effective and intuitive. Given an ${n}\times{n}$ entity table, for entity pair $(e_s, e_o)$, attending to its axial elements corresponds to attending to elements that are either $(e_s, e_i)$ or $(e_i, e_o)$. That is, if a two-hop relation $(e_s, e_o)$ can be dissected into a path $(e_s, e_i)$ and $(e_i, e_o)$, then the most informative neighbors for classifying $(e_s, e_o)$ are the one-hop candidates that share $e_{s}$ or $e_{o}$ with this entity pair. 
The axial attention is simply computed by self-attention along the height axis and the width axis, and each computation along the axes is followed by a residual connection. For the cell $(e_{s}, e_{o})$, we have:
\begin{equation}\small
\begin{split}
        r^{(s, o)}_{w}\!=\!r^{(s,o)}_{h}\!+\!
    \sum_{{p}\in{1..n}}{softmax_{p}}({q}_{(s, o)}^{T}k_{(s, p)})v_{(s, p)} \\
    r^{(s, o)}_{h}\!=\!g^{(s,o)}\!+\!
    \sum_{{p}\in{1...n}}{softmax_{p}}({q}_{(s, o)}^{T}k_{(p, o)})v_{(p, o)}
\end{split}
\end{equation}
where we denote query ${q}_{(i, j)} = W_{Q}g^{(i, j)}$, key ${k}_{(i, j)} = W_{K}g^{(i, j)}$, and value ${v}_{(i, j)} = W_{V}g^{(i, j)}$, which are all linear projections of the entity pair representation $g$ at position $(i, j)$. $W_{Q} \in \mathbb{R}^{{d}\times{d}}$, $W_{K} \in \mathbb{R}^{{d}\times{d}}$, and $W_{V} \in \mathbb{R}^{{d}\times{d}}$ are all learnable weight matrices. The output of the axial attention module is $r^{(s, o)}_{w} \in \mathbb{R}^d$. The $softmax_{p}$ function denotes a softmax function that applies to all possible $p = (i, j)$ positions. The formulation of this mechanism resembles \citet{wang2020axial}. However, our motivation is different, as \citet{wang2020axial} aim to use this mechanism to reduce the computational complexity of semantic segmentation, whereas our motivation is to attend to the one-hop neighbors for the two-hop relation triplets. 
 

\subsubsection{Adaptive Focal Loss}

 
Finally, we have a linear layer for predicting relations:
\begin{equation}
    {l}^{(s,o)} = \mathbf{W}_{l}r^{(s, o)}_{w} + b_{l}
\end{equation}
where ${l}^{(s,o)} \in \mathbb{R}^{c}$ denotes the output logits for all relations, $\mathbf{W}_{l} \in \mathbb{R}^{{d} \times {c}}$ is the weight matrix that maps the relation embedding to the logit of each class and $c$ is the number of classes. 

Our relation extraction problem is essentially a multi-label classification problem. Traditionally, binary cross-entropy (BCE) loss is used to tackle this problem. However, this method relies on a global probability threshold for inference. Recently \textbf{Adaptive Thresholding Loss} (\textbf{ATL}, \citealp{zhou2021document}) has been proposed for multi-label classification. Instead of using a global probability threshold for all examples, ATL introduced a special class $TH$ as the adaptive threshold value for each example. For each entity pair $(e_s, e_o)$, the classes whose logits are larger than the $TH$ class logit will be predicted as positive classes, and the rest will be predicted as negative classes. 

We propose \textbf{Adaptive Focal Loss} (\textbf{AFL}) as an enhancement to \textbf{ATL} for long-tail classes. Our loss consists of two parts, the first part is for positive classes and the second part is for negative classes. During training, the label space is divided into two subsets: positive class subset $\mathcal{P}_{T}$ and negative class subset $\mathcal{N}_{T}$. The positive class subset $\mathcal{P}_{T}$ contains the relations that exist in entity pair $(e_s, e_o)$, and if there is no relation between $(e_s, e_o)$, $\mathcal{P}_{T}$ is empty ($\mathcal{P}_{T}  = \varnothing$). The negative subset $\mathcal{N}_{T}$, on the other hand, contains the relation classes that do not belong to the positive classes, $\mathcal{N}_{T} = \mathcal{R} \setminus \mathcal{P}_{T}$. The probability of each positive class is computed as:
\begin{equation}
    P(r_{i}|e_{s}, e_{o}) = \frac{exp({{l}^{(s,o)}_{r_i})}}{ exp({l}^{(s,o)}_{r_i}) + 
    exp({l}^{(s,o)}_{TH})}
\end{equation}
where the logit of $r_{i}$ is ranked with the logit of threshold class $TH$ individually. This is different from the original ATL, where all positive logits are ranked together with a softmax function. For simplicity, $P(r_{i}|e_{s}, e_{o})$ is denoted as $P(r_{i})$ in this section, because we are only discussing $(e_s, e_o)$. For the negative classes, we use their logits to compute the probability of the $TH$ class:
\begin{equation}\small
    P(r_{TH}|e_{s}, e_{o}) = \frac{exp({l}^{(s,o)}_{r_{TH}})}{\sum_{{r_j} \in \mathcal{N}_{T} \cup \{TH\}} exp({l}^{(s,o)}_{r_j})} 
\end{equation}
Similarly, $P(r_{TH}|e_{s}, e_{o})$ is referred to as $P(r_{TH})$ in the remainder of this section. Since the distribution of the positive labels is highly imbalanced, we leverage the idea of focal loss \citep{lin2017focal} for balancing the logits of the positive classes. We have our loss function as:
\begin{equation}
\label{eq:re_loss}\small
    \mathcal{L}_{RE}\!=\!\sum_{r_i \in \mathcal{P}_{T}} (1\!-\!P(r_{i}))^{\gamma}log(P(r_{i}))\!+\!log(P(r_{TH}))
\end{equation}
where $\gamma$ is a hyper-parameter. Our loss is designed to focus more on the low-confidence classes. If $P(r_{i})$ is low, the loss contribution from the relevant class will be higher, which enables a better optimization for long-tail classes.  

\subsubsection{Knowledge Distillation for Distant Supervison}
In this section, we describe how we utilize the distantly supervised data in a more effective manner. The distantly supervised data included in the DocRed dataset \citep{yao2019docred} was obtained by performing entity linking on the Wikidata Knowledge Base~\citep{vrandevcic2014wikidata} and the Wikipedia data dump.  It is shown that pre-training from the distantly supervised data is beneficial for document-level relation extraction~\citep{xu2021entity}. However, prior work only adapts the distantly supervised data in a naive manner. The key challenge for the distant supervision adaptation is to overcome the differences between probability distributions of the distantly supervised data and the human annotated data. We compare two strategies for adapting the distantly supervised data.

    \paragraph{Naive Adaptation} Adopting from \citep{xu2021entity}, this method first pretrains the model with the distantly supervised data with the relation extraction loss $\mathcal{L}_{RE}$ (Eqn.~\ref{eq:re_loss}), and then the model is fine-tuned on the human-annotated data with the same objective. We denote this method as \textbf{Naive Adaptation (NA)}.
    
    \paragraph{Knowledge Distillation} To further utilize the annotated data, we use a relation classification model trained on the human-annotated data (\#Train in Table~\ref{tab:dataset-stats}) as the teacher model. The teacher model is used to generate soft labels on the distantly supervised data. Specifically, the distantly supervised data is fed into the teacher model and the predicted logits will be the soft labels used for training the student model. The student model has the same configuration as the teacher model, but is trained with two signals simultaneously. The first signal is the supervision from the hard labels of the distantly supervised data and the second is from the predicted soft labels. We denote the loss computed on the hard labels as $\mathcal{L}_{RE}$ and the knowledge distillation loss computed on the soft labels as $\mathcal{L}_{KD}$. We use mean squared error (MSE) as the knowledge distillation loss function:
    \begin{equation}
    \abovedisplayskip=2pt
    \belowdisplayskip=2pt
        \mathcal{L}_{KD} = MSE({l}^{(s,o)}_{S} ,{l}^{(s,o)}_{T})
    \end{equation}
    where ${l}^{(s,o)}_{S}$ denotes the predicted logits of the student model and ${l}^{(s,o)}_{T}$ is the prediction of the teacher model.
The student model is further fine-tuned with human-annotated data (\#Train in Table~\ref{tab:dataset-stats}) after it has been pre-trained on the distantly supervised data. 
    The overall loss of pre-training with distantly supervised data is computed as:
    \begin{equation}
    \abovedisplayskip=2pt
    \belowdisplayskip=2pt
        \mathcal{L} =  \mathcal{L}_{KD} + \mathcal{L}_{RE}
    \end{equation}
    We denote this method as \textbf{KD} in our main experimental results section. Besides the MSE loss, we also compare different adaptation methods, such as KL-Divergence, in section~\ref{sec:compare-adaptation}.

\section{Experiments}
\begin{table}[ht]
\centering
\resizebox{\columnwidth}{!}{
\begin{tabular}{lcc} 
\hline
Statistics                  & DocRED & HacRED  \\ 
\hline
\#  distant docs                   & 101,873   &   --   \\
\#  training docs                  & 3,053   & 6,231    \\
\#  dev docs                     & 1,000   & 1,500    \\
\#  test docs                    & 1,000   & 1,500    \\
\#  relations               & 97     & 27      \\
Avg \#  entities per doc    & 19.5   & 10.8    \\
Avg \# mentions per entity     & 1.4    & 1.2     \\
Avg \# relations per doc & 12.5   & 7.4     \\
\hline
\end{tabular}}

\caption{Dataset statistics of the DocRED and HacRED datasets.}
\label{tab:dataset-stats}
\end{table}
\subsection{Dataset Statistics}
We evaluated our model on two document-level relation extraction datasets -- the DocRED~\citep{yao2019docred} benchmark and the HacRED dataset~\citep{cheng-etal-2021-hacred}. DocRED is a crowd-sourced large-scale document-level relation extraction dataset. It contains 3,053/1,000/1,000 instances for training,
validation, and test, respectively. HacRED is a Chinese relation extraction dataset that focuses on the hard cases of relation extraction. It contains 27 hard relations and is split into 6,231/1,500/1,500 instances for training, validation, and test. However, the test set of HacRED is not released yet. In this paper, we only provide the results on its dev set.
\begin{table*}
\centering
\resizebox{0.7\textwidth}{!}{
\begin{tabular}{lcccc} 
\hline
& \multicolumn{2}{c}{Dev}  & \multicolumn{2}{c}{Test}   \\ 
\textit{w/o Distant Supervision}& Ign\_F1 & F1 & Ign\_F1 & F1  \\ 
\hline

\textbf{Two-stage-B-b}                & 56.67        & 58.83   & 56.47         & 58.69     \\
\textbf{ATLOP-B-b}       & 59.22$\pm{0.15}$        & 61.09$\pm{0.16}$   & 59.31         & 61.30      \\
\textbf{SIRE-B-b}       & 59.82        & 61.60   & \textbf{60.18}         & 62.05      \\
\textbf{DocuNet-B-b}       & 59.86$\pm{0.13}$        & 61.83$\pm{0.19}$   & 59.93         & 61.86      \\
\hline
\textbf{Ours-B-b}      & \textbf{60.08}$\pm{0.11}$       & \textbf{62.03}$\pm{0.18}$   & 60.04         & \textbf{62.08}      \\
\hline
\textbf{Coref-Rb-l}       & 57.35        & 59.43   & 57.9          & 60.25     \\
\textbf{SSAN-Rb-l}        & 59.40         & 61.42   & 60.25         & 62.08     \\
\textbf{GAIN-B-l}     & 60.87        & 63.09   & 60.31         & 62.76     \\
\textbf{ATLOP-Rb-l}       & 61.32$\pm{0.14}$        & 63.18$\pm{0.19}$   & 61.39         & 63.40      \\
\textbf{DocuNet-Rb-l}            & \textbf{62.23}$\pm{0.12}$        & 64.12$\pm{0.14}$   & 62.39         & \textbf{64.55} \\
\textbf{DocuNet-Rb-l}$^*$             & 61.56$\pm{0.14}$        & 63.58$\pm{0.17}$   & 61.79         & 63.73     \\
\hline
\textbf{Ours-Rb-l}                & 62.16$\pm{0.10}$        & \textbf{64.19}$\pm{0.16}$   & \textbf{62.57}         & 64.28     \\
\hline
\hline
\textit{with Distant Supervision} & Ign\_F1 & F1 & Ign\_F1 & F1    \\
\hline
 \textbf{ATLOP-NA-Rb-l$^*$}         & 63.41$\pm{0.15}$        & 65.33$\pm{0.18}$   & 63.54         & 65.47     \\
  \textbf{DocuNet-NA-Rb-l$^*$}         & 63.26$\pm{0.17}$       & 65.21$\pm{0.19}$  & 63.29         & 65.44     \\
 \textbf{SSAN-NA-Rb-l}         & 63.76        & 65.69   & 63.78         & 65.92     \\ 
 \hline
   \textbf{Ours-NA-B-b}    & 62.18$\pm{0.12}$	& 64.17$\pm{0.16}$	& 61.77	& 64.12     \\
 \textbf{Ours-KD-B-b}    & 62.62$\pm{0.16}$	& 64.81$\pm{0.13}$	& 62.56  &	 64.76     \\
\textbf{Ours-NA-Rb-l}         & 63.38$\pm{0.11}$        & 65.64$\pm{0.17}$   & 63.63         & 65.71     \\ 
\textbf{Ours-KD-Rb-l}    & \textbf{65.27}$\pm{0.09}$        & \textbf{67.12}$\pm{0.14}$   & \textbf{65.24}         & \textbf{67.28}     \\
\hline
\hline
\end{tabular}}

\caption{Experimental results for the DocRED dataset. The reported metrics are F1 score and Ign\_F1. We report the average of five random runs for the development set and the best checkpoint is used for the leaderboard submission for the test results. Results with $^*$ are obtained by our reproduction.}
\label{tab:docred}
\end{table*}

\subsection{Implementation Details}
We implemented our model with the PyTorch version of the Huggingface Transformers~\citep{wolf2020transformers}. For experiments on DocRED, we experimented with Roberta-large~\citep{liu2019roberta} and Bert-base~\citep{devlin-etal-2019-bert} as our document encoder respectively. For experiments on HacRED, we use XLM-R base~\citep{conneau-etal-2020-unsupervised} as the document encoder. AdamW~\citep{loshchilov2018decoupled} is used as the optimizer. At the knowledge distillation stage, we trained the model with the learning rate set to 1e-5 for 2 epochs. Warmup is applied on the initial 6\% steps. The dropout rates between transformer layers are set to 0.1 and the maximum gradient norm is clipped at 1.0. During the fine-tuning stage, the learning rate is set to 1e-6 and we train the model for 10 epochs. We performed grid search for $\gamma \in [0, 0.5, 1.0, 1.5, 2.0]$ and set it to 0.5. Our model is trained on a single NVIDIA V100 GPU with 32 GB memory. The main evaluation metrics are Ign\_F1 and F1 score following \citet{yao2019docred}, where Ign\_F1 refers to the F1 score that ignores the triples that appear in the annotated training data.
\subsection{Compared Methods}
We denote Bert-base and Bert-large encoders as \textbf{B-b} and \textbf{B-l}. The Roberta-large model is denoted as \textbf{Rb-l}. We compare our model with the state-of-the-art systems on the DocRED leaderboard as well as strong baselines by our own implementation. They are the following models: \citet{wang2019fine} has proposed to fine-tune BERT for document-level RE with a two-step process (\textbf{Two-stage-B-b}). The Bert model needs to classify whether the two entities have relation and then classify their relation if the first step is positive. The \textbf{Coref-Rb-l} \citep{ye2020coreferential} uses a co-reference module to aggregate the mention representations of the same entity. The \textbf{SSAN}~\citep{xu2021entity} model utilizes co-occurrence information between entity mentions, leverages distantly supervised data for pretraining, and achieves the state of the art on the DocRED leaderboard. Since their best model \textbf{SSAN-Adapt} is equivalent to naive adaptation in our work, we denote it as \textbf{SSAN-NA-Rb-l} in our experiments. The \textbf{GAIN}~\citep{zeng2020double} model adds a graph neural network on top of a pre-trained language model, constructs a document-level graph for each example, and uses the graphical structure to extract relations. \textbf{SIRE}~\citep{zeng2021sire} uses two encoders for different types of relation --- a sentence-level encoder to extract intra-sentence relations and a document encoder to extract inter-sentence relations. \textbf{ATLOP}~\citep{zhou2021document} is purely based on the transformer architecture and a novel adaptive thresholding loss to deal with the multi-label problem for DocRE. Besides, it also fuses the contextual information with the aggregated attention weights for each entity. The \textbf{DocuNet}~\citep{zhang2021document} model treats the relation extraction task in a similar way as semantic segmentation in computer vision. We also conducted an experiment that pretrained the \textbf{ATLOP-Rb-l} model with distantly supervised data, as this model is the best model by our reproduction.

\subsection{Main Results}
Our main results for the DocRED dataset are shown in Table~\ref{tab:docred}. Knowledge distillation is able to significantly improve the performance of our model. \textbf{Ours-KD-Rb-l} achieves the best single-model performance of 67.28 test F1. Our best model significantly ourperforms the previous state of the art \textbf{SSAN-NA-Rb-l} by 1.36 on test F1 and 1.46 on test Ign\_F1. As of 11th Nov 2021, our best model achieves the highest scores on the DocRED leaderboard.


\begin{table}[ht]
\centering

\begin{tabular}{llll} 
\hline
~     & P     & R     & F1     \\ 
\hline
\textbf{GAIN}$^*$   & 73.38 & \textbf{80.07} & 76.09  \\
\textbf{ATLOP}$^*$ & 76.97 & 78.29 & 77.63  \\
\textbf{Ours}  & \textbf{78.53} & 78.96 & \textbf{78.75}  \\
\hline
\end{tabular}
\setlength{\abovecaptionskip}{2pt}
\caption{Experimental results on HacRED dev set. Results with $^*$ are implemented by us. All experiments used XLM-R-base as the encoder.}
\label{tab:hacred}
\end{table}

The experiment results for the HacRED dataset are shown in Table~\ref{tab:hacred}. The main difference of our method with the ATLOP baseline is the Adaptive Focal Loss and the Axial Attention Module. Our proposed method is able to exceed the ATLOP baseline by 1.12 F1. Besides the performance of the models, it is worth noting that for each method, the absolute performance of HacRED is significantly higher than its performance on DocRED. This is counter-intuitive as HacRED focuses on the hard relations whereas DocRED is more general. This can be caused by the following: 1) The human annotated training instances of the HacRED dataset are significantly more than DocRED, leading to better generalization performance. 2) Even though HacRED claims it focuses on the hard cases for relation extraction, it only has 27 classes, and the relation type distribution within the HacRED dataset is more balanced. 

\begin{table}[ht]
\centering
\resizebox{1.0\columnwidth}{!}{
\begin{tabular}{lccc} 
\hline
~                       & Frequent
   & Long-tail
   & Overall
   \\ 
~                       & F1
   & F1
   & F1
   \\ 
\hline
\textbf{ATLOP-Rb-l}                   & 70.93         & 50.01            & 63.12         \\
\textbf{Ours-Rb-l}                   & \textbf{71.26}         & \textbf{51.97}            & \textbf{64.19}         \\
 \textit{w/o} \textbf{Axial}          & 70.86	& 50.77	 &   63.56         \\
 \textit{w/o} \textbf{AFL} \textit{w} \textbf{ATL} & 70.94	& 50.86  &	63.67        \\
\hline
\textit{With Distant Supervision}                   &          &           &         \\
\hline
\textbf{ATLOP-NA-Rb-l}            & 73.26         & 52.39            & 65.33         \\
\textbf{Ours-KD-Rb-l}                    & \textbf{74.15}         & \textbf{56.51}            & \textbf{67.12}         \\
 \textit{w/o} \textbf{Axial}          & 73.52         & 54.96            & 66.36         \\
 
 \textit{w/o} \textbf{AFL} \textit{w} \textbf{ATL} & 73.50          & 54.73            & 66.23         \\
\hline
\end{tabular}}
\setlength{\abovecaptionskip}{0pt}
\caption{Experiment results for frequent and long-tail type relations. Frequent types refer to the most popular 10 relation types, and long-tail relations refer to the rest of the 86 relations. }
\label{tab:long-tail-results}
\end{table}

\subsection{Ablation Study}
We first separate our label space into two subsets. The first subset consists of the 10 most frequent labels, accounting for 59.4\% of the positive relations in the training data. The second subset is denoted as the long-tail labels, which includes the rest of the 86 relations (the total label space is 97 and there is one $TH$ class). Since our Adaptive Focal loss function is mainly designed for improving the performance on the less frequent classes, we show the ablation study by frequent and long-tail classes in Table~\ref{tab:long-tail-results}. When we change the AFL loss to conventional Adaptive Thresholding Loss~\citep{zhou2021document}, the overall performance with KD drops by 0.89 F1, and the F1 score for the frequent labels only drops by 0.65. Meanwhile, the long-tail labels' F1 drops by 1.78, which is significantly higher than the drop in overall performance and frequent performance. This indicates that our Adaptive Focal Loss is able to balance the weight of the frequent classes and infrequent classes. The axial attention module is also more beneficial for the long-tail classes than the frequent classes, which shows that our model's performance on the frequent classes is saturated.

\begin{table}[ht]
\centering

\begin{tabular}{lccc} 
\hline
~                 & P     & R     & Infer-F1  \\ 
\hline
\textbf{GAIN-B-b}           & 38.71 & 59.45 & 46.89           \\
\textbf{Ours-Rb-l} & \textbf{42.15} & \textbf{61.56} & \textbf{50.04}           \\
 \textit{w/o} \textbf{Axial}    & 40.26 & 60.60  & 48.37           \\
\hline
\end{tabular}

\caption{Ablation study for the Infer-F1 relation triples on the development set of DocRED.}
\label{tab:infer-result}

\end{table}

We also provide an ablation study on the multi-hop relations in Table~\ref{tab:infer-result}. We use the same evaluation method for multi-hop relations as \citet{zeng2020double}. This evaluation method ignores all the one-hop relation triples. Our axial attention module effectively improves Infer-F1 by 1.67, while its improvement for overall performance is only 0.63. 

\subsection{Comparison of Adaptation Methods}
\label{sec:compare-adaptation}

\begin{table}[ht]
\centering

\begin{tabular}{lrr} 
\hline
\textit{Distant Adaptation} & \multicolumn{1}{l}{Ign\_F1} & \multicolumn{1}{l}{F1}  \\ 
\hline
\textbf{NA}                 & 52.29                       & 54.67                             \\            
\textbf{KD$_{KL}$}             & 53.89                       & 56.97                             \\
\textbf{KD$_{MSE}$}            & \textbf{55.28}                       & \textbf{57.74}                             \\
\hline
\textit{Continue-trained} & \multicolumn{1}{l}{Ign\_F1} & \multicolumn{1}{l}{F1}  \\ 
\hline
\textbf{NA}                 & 63.38                       & 65.64                              \\            
\textbf{KD$_{KL}$}             & 64.42                       & 66.24                             \\
\textbf{KD$_{MSE}$}            & \textbf{65.27}                       & \textbf{67.12}                             \\
\hline
\end{tabular}
\setlength{\abovecaptionskip}{0pt}
\caption{Development set performance of different knowledge adaptation methods for DocRED.}
\label{tab:distant-results}
\end{table}

In this section, we directly compare the knowledge adaptation methods on the development set of DocRED (Table~\ref{tab:distant-results}). We mainly compare three methods for adaptation: 1) Naive Adaptation~(\textbf{NA}), 2) \textbf{KD$_{KL}$} knowledge distillation with the KL divergence loss and 3) \textbf{KD$_{MSE}$} with mean squared error loss. The adaptation performance on the development set is positively correlated with the performance of downstream fine-tuning. In the distant adaptation setting, our best method \textbf{KD$_{MSE}$} is able to outperform \textbf{NA} by 3.07 F1 and \textbf{KD$_{KL}$} by 0.77 F1. Similar performance differences are observed in the continue-trained setting.

\section{Error Analysis}
\label{sec:error-analysis}

Even though our final model significantly outperforms the previous state of the art on the DocRED leaderboard, the absolute performance of our model still does not match human performance. In this section, we provide a detailed error analysis of our model on the development set of DocRED.

\begin{table}[ht]
\newcommand{\multirot}[1]{\multirow{4}{*}[1.0ex]{\rotcell{\rlap{#1}}}}
\centering
\def\arraystretch{1.0}%
\resizebox{\columnwidth}{!}{
\begin{tabular}{l|l|l|l|}
\multicolumn{1}{l}{}          & \multicolumn{1}{l}{} & \multicolumn{2}{c}{\textbf{Ground Truth}}       \\ 
\cline{2-4}
\multirow{4}{*}{\rotcell{\clap{\textbf{Predictions}}}}&                      & $r \in$ \textbf{R} & \textbf{NR}                          \\ 
\cline{2-4}
                                &  \multirow{2}{*}{$r \in $ \textbf{R} }        & \textbf{C}: 8,273 (51.4\%)   & \multirow{2}{*}{\textbf{MR}: 3,814 (23.7\%)}  \\ 
\cline{3-3}
                                &            & \textbf{W}: 242 (1.5\%)     &                             \\ 
\cline{2-4}
                                & \textbf{NR}                   & \textbf{MS}: 3,761 (23.4\%)   & 380,703                    \\
\cline{2-4}
\end{tabular}}
\caption{Statistics of our error distribution. The final evaluation score is evaluated on $r \in $ \textbf{R} triples, hence the correct predictions of \textbf{NR} are ignored when calculating the final scores.}
\label{tab:error-types}
\end{table}

We first construct the union of our model's predictions and the ground truth triples (without \textbf{NR} label). Then, we categorize the union into four categories: (1) \textbf{Correct (C)}, where prediction triples are in the ground truth. (2) \textbf{Wrong (W)}, where the predicted head entity and tail entity are in the ground truth but the predicted relation is wrong. (3) \textbf{Missed (MS)}, where the model predicts no relation for a pair of head entity and tail entity with some relation in the ground truth. (4) \textbf{More (MR)}, where the model predicts an extraneous relation for a pair of head entity and tail entity not related in the ground truth. From Table~\ref{tab:error-types}, we observe that the error percentage of the \textbf{W} category is very small. This indicates that for a pair of head entity and tail entity with some relation in the ground truth, and when our model predicts that there is a relation between these two entities, it is able to predict the correct relation rather accurately. However, we observe that most of our errors are under the \textbf{MR} and \textbf{MS} categories, and their counts are about the same. To better understand the performance bottleneck of the document-level RE task, we evaluate our model on a simplified subtask (Table~\ref{tab:sub-task}). This subtask is binary classification, i.e., to determine whether two entities are related or not, and it is denoted as \textbf{Binary Labels}. In this subtask, we only care about predicting correctly that there is some relation between a head entity and a tail entity, but not what the exact relation is among the 97 relation classes. The performance on this simplified task is 68.64 F1 score, which is only marginally higher than the original F1 score of 67.12. This may be due to incomplete annotation of the two document-level relation extraction datasets, and we will illustrate this hypothesis in Figure~\ref{fig:example-output}.   


\begin{table}[ht]
\centering

\begin{tabular}{lrrr} 
\hline
                     & \multicolumn{1}{l}{P} & \multicolumn{1}{l}{R} & \multicolumn{1}{l}{F1}  \\ 
\hline
\textbf{Binary Labels}         & 68.51                 & 68.78                 & 68.64                   \\
\textbf{Original Labels}              & 67.10                  & 67.13                 & 67.12                   \\

\hline
\end{tabular}
\setlength{\abovecaptionskip}{2pt}
\caption{Performance breakdown on the DocRED dev set.}
\label{tab:sub-task}
\end{table}


\begin{figure}[ht]
\centering
\small
\begin{tabular}{p{1.0\columnwidth}}

``\textbf{Eivind Bolle} ( \textbf{13 October 1923} – \textbf{10 June 2012} ) was a \textbf{Norwegian} politician for the \textbf{Labour Party}. He was born in \textbf{Hol}. He was elected to the \textbf{Norwegian Parliament} from \textbf{Nordland} in 1973. ...  On the local level he was a member of \textbf{Hol} municipality council from \textbf{1959} to \textbf{1963} , and later in Hol 's successor municipality \textbf{Vestvågøy}. He served as mayor from \textbf{1971} to \textbf{1973} , during which term he was also a member of \textbf{Nordland} county council ..."~ ~ ~~  \\ \\
\textbf{More}: (\textbf{Nordland}, \textit{country}, \textbf{Norwegian}), (\textbf{Vestvågøy}, \textit{country}, \textbf{Norwegian}),...\\ \\
\textbf{Correct}: (\textbf{Labour Party}, \textit{country}, \textbf{Norwegian}), (\textbf{Hol}, \textit{country}, \textbf{Norwegian}),...\\

\end{tabular}
\setlength{\abovecaptionskip}{1pt}

\caption{Example output of our model on the DocRED dev set.}
\label{fig:example-output}
\end{figure}

In Figure~\ref{fig:example-output}, we show an example document from the dev set of DocRED and its predictions. We observe that many triples in the \textbf{MR} category are factually correct. That is, some of the pairs of entities are truly related but are labeled as \textbf{NR} throughout the dataset. For instance, from the ground truth, we can see that \textbf{Labour Party} and \textbf{Hol} are all entities from \textit{country} \textbf{Norway}. Similarly, \textbf{Nordland} and \textbf{Vestvågøy} are all in \textbf{Norway}, but their relations with \textbf{Norway} are not present in the ground truth triples. Therefore, when our model predicts these triples, its performance would be unfairly penalized during evaluation. This observation indicates that there are some incomplete annotations in the DocRED dataset. However, this is not the focus of this paper and we would like to leave this as future work.


\section{Related Work}
Early works on relation extraction mainly focused on sentence-level RE (\citealp{zhang-etal-2017-position}; \citealp{baldini-soares-etal-2019-matching}; \citealp{peng2020learning}). However, prior works have shown that a large number of relations can only be extracted from multiple sentences (\citealp{verga-etal-2018-simultaneously}; \citealp{yao2019docred}; \citealp{cheng-etal-2021-hacred}). Various methods have been proposed to tackle document-level relation extraction (DocRE). Graph neural networks~(GNNs; \citealp{scarselli2008graph}) have been widely used for the DocRE task. \citet{quirk-poon-2017-distant} used words as nodes and dependency information as edges to construct document-level graphs. This graph will be used to extract features for each entity pair. Later works extended this idea by applying different GNN architectures (\citealp{peng2017cross}; \citealp{verga-etal-2018-simultaneously}; \citealp{christopoulou-etal-2019-connecting}; \citealp{nan-etal-2020-reasoning}; \citealp{zhang-etal-2018-graph}; \citealp{zeng2020double}). In particular, \citet{nan-etal-2020-reasoning} proposed the latent stucture refinement (LSR) model, which used structured attention to induce the document-level graph. \citet{zeng2020double} constructed the document-level graph by entity-mention nodes and sentence edges. Besides the graph-based methods, transformer-only architectures have also proven to be highly effective for the DocRE task (\citealp{tang2020hin}; \citealp{zhou2021document}). Specifically, \citet{zhou2021document} proposed adaptive thresholding loss to tackle the multi-label classification problem in DocRE.

On the other hand, learning from distant supervision is another important problem for relation extraction. \citet{qin-etal-2018-dsgan} used generative adversarial training for selecting informative examples and \citet{feng2018reinforcement} used reinforcement learning to achieve the same goal. However, there are no existing works that jointly learn from annotated data and distant data. To this end, this paper is the first to overcome the differences between the human annotated and distantly supervised data. Moreover, this paper also tackles the under-explored class imbalance problem and the two-hop logical reasoning problem with novel solutions to the shortcomings of existing approaches. 
\section{Conclusions}
In this paper, we have proposed a novel framework for document-level relation extraction, based on knowledge distillation, axial attention, and adaptive focal loss. Our proposed method is able to significantly outperform the previous state of the art on the DocRED leaderboard. Besides, we also conducted a thorough ablation study and error analysis to identify the bottleneck of the document-level relation extraction task.

\section{Acknowledgements}
We would like to thank the anonymous reviewers for their insightful feedback and comments.

\bibliographystyle{acl_natbib}
\bibliography{custom}

\begin{thebibliography}{32}
\expandafter\ifx\csname natexlab\endcsname\relax\def\natexlab#1{#1}\fi

\bibitem[{Baldini~Soares et~al.(2019)Baldini~Soares, FitzGerald, Ling, and
  Kwiatkowski}]{baldini-soares-etal-2019-matching}
Livio Baldini~Soares, Nicholas FitzGerald, Jeffrey Ling, and Tom Kwiatkowski.
  2019.
\newblock \href {https://aclanthology.org/P19-1279} {Matching the blanks:
  Distributional similarity for relation learning}.
\newblock In \emph{Proceedings of ACL}.

\bibitem[{Cheng et~al.(2021)Cheng, Liu, Qu, Zhao, Liang, Wang, Huai, Yuan, and
  Xiao}]{cheng-etal-2021-hacred}
Qiao Cheng, Juntao Liu, Xiaoye Qu, Jin Zhao, Jiaqing Liang, Zhefeng Wang,
  Baoxing Huai, Nicholas~Jing Yuan, and Yanghua Xiao. 2021.
\newblock \href {https://aclanthology.org/2021.findings-acl.249} {{H}ac{RED}: A
  large-scale relation extraction dataset toward hard cases in practical
  applications}.
\newblock In \emph{Findings of ACL}.

\bibitem[{Christopoulou et~al.(2019)Christopoulou, Miwa, and
  Ananiadou}]{christopoulou-etal-2019-connecting}
Fenia Christopoulou, Makoto Miwa, and Sophia Ananiadou. 2019.
\newblock \href {https://aclanthology.org/D19-1498} {Connecting the dots:
  Document-level neural relation extraction with edge-oriented graphs}.
\newblock In \emph{Proceedings of EMNLP}.

\bibitem[{Conneau et~al.(2020)Conneau, Khandelwal, Goyal, Chaudhary, Wenzek,
  Guzm{\'a}n, Grave, Ott, Zettlemoyer, and
  Stoyanov}]{conneau-etal-2020-unsupervised}
Alexis Conneau, Kartikay Khandelwal, Naman Goyal, Vishrav Chaudhary, Guillaume
  Wenzek, Francisco Guzm{\'a}n, Edouard Grave, Myle Ott, Luke Zettlemoyer, and
  Veselin Stoyanov. 2020.
\newblock \href {https://doi.org/10.18653/v1/2020.acl-main.747} {Unsupervised
  cross-lingual representation learning at scale}.
\newblock In \emph{Proceedings of ACL}.

\bibitem[{Devlin et~al.(2019)Devlin, Chang, Lee, and
  Toutanova}]{devlin-etal-2019-bert}
Jacob Devlin, Ming-Wei Chang, Kenton Lee, and Kristina Toutanova. 2019.
\newblock \href {https://aclanthology.org/N19-1423} {{BERT}: Pre-training of
  deep bidirectional transformers for language understanding}.
\newblock In \emph{Proceedings of NAACL}.

\bibitem[{Feng et~al.(2018)Feng, Huang, Zhao, Yang, and
  Zhu}]{feng2018reinforcement}
Jun Feng, Minlie Huang, Li~Zhao, Yang Yang, and Xiaoyan Zhu. 2018.
\newblock \href
  {https://www.aaai.org/ocs/index.php/AAAI/AAAI18/paper/download/17151/16140}
  {Reinforcement learning for relation classification from noisy data}.
\newblock In \emph{Proceedings of AAAI}.

\bibitem[{Li et~al.(2021)Li, Xu, Li, Fei, Ren, and Ji}]{li2021mrn}
Jingye Li, Kang Xu, Fei Li, Hao Fei, Yafeng Ren, and Donghong Ji. 2021.
\newblock \href {https://aclanthology.org/2021.findings-acl.117.pdf} {Mrn: A
  locally and globally mention-based reasoning network for document-level
  relation extraction}.
\newblock In \emph{Findings of ACL}.

\bibitem[{Lin et~al.(2017)Lin, Goyal, Girshick, He, and
  Doll{\'a}r}]{lin2017focal}
Tsung-Yi Lin, Priya Goyal, Ross Girshick, Kaiming He, and Piotr Doll{\'a}r.
  2017.
\newblock \href
  {https://openaccess.thecvf.com/content_ICCV_2017/papers/Lin_Focal_Loss_for_ICCV_2017_paper.pdf}
  {Focal loss for dense object detection}.
\newblock In \emph{Proceedings of ICCV}.

\bibitem[{Liu et~al.(2019)Liu, Ott, Goyal, Du, Joshi, Chen, Levy, Lewis,
  Zettlemoyer, and Stoyanov}]{liu2019roberta}
Yinhan Liu, Myle Ott, Naman Goyal, Jingfei Du, Mandar Joshi, Danqi Chen, Omer
  Levy, Mike Lewis, Luke Zettlemoyer, and Veselin Stoyanov. 2019.
\newblock \href {https://arxiv.org/abs/1907.11692} {Roberta: A robustly
  optimized bert pretraining approach}.
\newblock \emph{arXiv preprint arXiv:1907.11692}.

\bibitem[{Loshchilov and Hutter(2019)}]{loshchilov2018decoupled}
Ilya Loshchilov and Frank Hutter. 2019.
\newblock \href {https://openreview.net/forum?id=Bkg6RiCqY7} {Decoupled weight
  decay regularization}.
\newblock In \emph{Proceedings of ICLR}.

\bibitem[{Nan et~al.(2020)Nan, Guo, Sekulic, and Lu}]{nan-etal-2020-reasoning}
Guoshun Nan, Zhijiang Guo, Ivan Sekulic, and Wei Lu. 2020.
\newblock \href {https://aclanthology.org/2020.acl-main.141} {Reasoning with
  latent structure refinement for document-level relation extraction}.
\newblock In \emph{Proceedings of ACL}.

\bibitem[{Peng et~al.(2020)Peng, Gao, Han, Lin, Li, Liu, Sun, and
  Zhou}]{peng2020learning}
Hao Peng, Tianyu Gao, Xu~Han, Yankai Lin, Peng Li, Zhiyuan Liu, Maosong Sun,
  and Jie Zhou. 2020.
\newblock \href {https://aclanthology.org/2020.emnlp-main.298.pdf} {Learning
  from context or names? {An} empirical study on neural relation extraction}.
\newblock In \emph{Proceedings of EMNLP}.

\bibitem[{Peng et~al.(2017)Peng, Poon, Quirk, Toutanova, and
  Yih}]{peng2017cross}
Nanyun Peng, Hoifung Poon, Chris Quirk, Kristina Toutanova, and Wen-tau Yih.
  2017.
\newblock \href {https://aclanthology.org/Q17-1008/} {Cross-sentence n-ary
  relation extraction with graph lstms}.
\newblock \emph{Transactions of the Association for Computational Linguistics}.

\bibitem[{Qin et~al.(2018)Qin, Xu, and Wang}]{qin-etal-2018-dsgan}
Pengda Qin, Weiran Xu, and William~Yang Wang. 2018.
\newblock \href {https://aclanthology.org/P18-1046} {{DSGAN}: Generative
  adversarial training for distant supervision relation extraction}.
\newblock In \emph{Proceedings of ACL}.

\bibitem[{Quirk and Poon(2017)}]{quirk-poon-2017-distant}
Chris Quirk and Hoifung Poon. 2017.
\newblock \href {https://aclanthology.org/E17-1110} {Distant supervision for
  relation extraction beyond the sentence boundary}.
\newblock In \emph{Proceedings of EACL}.

\bibitem[{Scarselli et~al.(2008)Scarselli, Gori, Tsoi, Hagenbuchner, and
  Monfardini}]{scarselli2008graph}
Franco Scarselli, Marco Gori, Ah~Chung Tsoi, Markus Hagenbuchner, and Gabriele
  Monfardini. 2008.
\newblock \href {https://ieeexplore.ieee.org/document/4700287} {The graph
  neural network model}.
\newblock \emph{IEEE Transactions on Neural Networks}.

\bibitem[{Tang et~al.(2020)Tang, Cao, Zhang, Cao, Fang, Wang, and
  Yin}]{tang2020hin}
Hengzhu Tang, Yanan Cao, Zhenyu Zhang, Jiangxia Cao, Fang Fang, Shi Wang, and
  Pengfei Yin. 2020.
\newblock {HIN:} hierarchical inference network for document-level relation
  extraction.
\newblock In \emph{Proceedings of KDD}.

\bibitem[{Vaswani et~al.(2017)Vaswani, Shazeer, Parmar, Uszkoreit, Jones,
  Gomez, Kaiser, and Polosukhin}]{vaswani2017attention}
Ashish Vaswani, Noam Shazeer, Niki Parmar, Jakob Uszkoreit, Llion Jones,
  Aidan~N Gomez, {\L}ukasz Kaiser, and Illia Polosukhin. 2017.
\newblock \href
  {https://proceedings.neurips.cc/paper/2017/file/3f5ee243547dee91fbd053c1c4a845aa-Paper.pdf}
  {Attention is all you need}.
\newblock In \emph{Proceedings of NIPS}.

\bibitem[{Verga et~al.(2018)Verga, Strubell, and
  McCallum}]{verga-etal-2018-simultaneously}
Patrick Verga, Emma Strubell, and Andrew McCallum. 2018.
\newblock \href {https://aclanthology.org/N18-1080} {Simultaneously
  self-attending to all mentions for full-abstract biological relation
  extraction}.
\newblock In \emph{Proceedings of NAACL}.

\bibitem[{Vrande{\v{c}}i{\'c} and Kr{\"o}tzsch(2014)}]{vrandevcic2014wikidata}
Denny Vrande{\v{c}}i{\'c} and Markus Kr{\"o}tzsch. 2014.
\newblock \href
  {https://cacm.acm.org/magazines/2014/10/178785-wikidata/fulltext} {Wikidata:
  a free collaborative knowledgebase}.
\newblock \emph{Communications of the ACM}.

\bibitem[{Wang et~al.(2019)Wang, Focke, Sylvester, Mishra, and
  Wang}]{wang2019fine}
Hong Wang, Christfried Focke, Rob Sylvester, Nilesh Mishra, and William Wang.
  2019.
\newblock \href {https://arxiv.org/abs/1909.11898} {Fine-tune {BERT} for
  {DocRED} with two-step process}.
\newblock \emph{arXiv preprint arXiv:1909.11898}.

\bibitem[{Wang et~al.(2020)Wang, Zhu, Green, Adam, Yuille, and
  Chen}]{wang2020axial}
Huiyu Wang, Yukun Zhu, Bradley Green, Hartwig Adam, Alan Yuille, and
  Liang-Chieh Chen. 2020.
\newblock \href
  {http://www.ecva.net/papers/eccv_2020/papers_ECCV/papers/123490103.pdf}
  {Axial-deeplab: Stand-alone axial-attention for panoptic segmentation}.
\newblock In \emph{Proceedings of ECCV}.

\bibitem[{Wolf et~al.(2020)Wolf, Chaumond, Debut, Sanh, Delangue, Moi, Cistac,
  Funtowicz, Davison, Shleifer et~al.}]{wolf2020transformers}
Thomas Wolf, Julien Chaumond, Lysandre Debut, Victor Sanh, Clement Delangue,
  Anthony Moi, Pierric Cistac, Morgan Funtowicz, Joe Davison, Sam Shleifer,
  et~al. 2020.
\newblock \href {https://aclanthology.org/2020.emnlp-demos.6} {Transformers:
  State-of-the-art natural language processing}.
\newblock In \emph{Proceedings of EMNLP: System Demonstrations}.

\bibitem[{Xu et~al.(2021)Xu, Wang, Lyu, Zhu, and Mao}]{xu2021entity}
Benfeng Xu, Quan Wang, Yajuan Lyu, Yong Zhu, and Zhendong Mao. 2021.
\newblock \href {https://ojs.aaai.org/index.php/AAAI/article/view/17665}
  {Entity structure within and throughout: Modeling mention dependencies for
  document-level relation extraction}.
\newblock In \emph{Proceedings of AAAI}.

\bibitem[{Yao et~al.(2019)Yao, Ye, Li, Han, Lin, Liu, Liu, Huang, Zhou, and
  Sun}]{yao2019docred}
Yuan Yao, Deming Ye, Peng Li, Xu~Han, Yankai Lin, Zhenghao Liu, Zhiyuan Liu,
  Lixin Huang, Jie Zhou, and Maosong Sun. 2019.
\newblock \href {https://aclanthology.org/P19-1074} {{DocRED:} a large-scale
  document-level relation extraction dataset}.
\newblock In \emph{Proceedings of ACL}.

\bibitem[{Ye et~al.(2020)Ye, Lin, Du, Liu, Li, Sun, and
  Liu}]{ye2020coreferential}
Deming Ye, Yankai Lin, Jiaju Du, Zhenghao Liu, Peng Li, Maosong Sun, and
  Zhiyuan Liu. 2020.
\newblock \href {https://aclanthology.org/2020.emnlp-main.582/} {Coreferential
  reasoning learning for language representation}.
\newblock In \emph{Proceedings of EMNLP}.

\bibitem[{Zeng et~al.(2021)Zeng, Wu, and Chang}]{zeng2021sire}
Shuang Zeng, Yuting Wu, and Baobao Chang. 2021.
\newblock \href {https://aclanthology.org/2021.findings-acl.47} {{SIRE}:
  Separate intra- and inter-sentential reasoning for document-level relation
  extraction}.
\newblock In \emph{Findings of ACL}.

\bibitem[{Zeng et~al.(2020)Zeng, Xu, Chang, and Li}]{zeng2020double}
Shuang Zeng, Runxin Xu, Baobao Chang, and Lei Li. 2020.
\newblock \href {https://aclanthology.org/2020.emnlp-main.127/} {Double graph
  based reasoning for document-level relation extraction}.
\newblock In \emph{Proceedings of EMNLP}.

\bibitem[{Zhang et~al.(2021)Zhang, Chen, Xie, Deng, Tan, Chen, Huang, Si, and
  Chen}]{zhang2021document}
Ningyu Zhang, Xiang Chen, Xin Xie, Shumin Deng, Chuanqi Tan, Mosha Chen, Fei
  Huang, Luo Si, and Huajun Chen. 2021.
\newblock \href {https://doi.org/10.24963/ijcai.2021/551} {Document-level
  relation extraction as semantic segmentation}.
\newblock In \emph{Proceedings of IJCAI}.

\bibitem[{Zhang et~al.(2018)Zhang, Qi, and Manning}]{zhang-etal-2018-graph}
Yuhao Zhang, Peng Qi, and Christopher~D. Manning. 2018.
\newblock \href {https://aclanthology.org/D18-1244} {Graph convolution over
  pruned dependency trees improves relation extraction}.
\newblock In \emph{Proceedings of EMNLP}.

\bibitem[{Zhang et~al.(2017)Zhang, Zhong, Chen, Angeli, and
  Manning}]{zhang-etal-2017-position}
Yuhao Zhang, Victor Zhong, Danqi Chen, Gabor Angeli, and Christopher~D.
  Manning. 2017.
\newblock \href {https://aclanthology.org/D17-1004} {Position-aware attention
  and supervised data improve slot filling}.
\newblock In \emph{Proceedings of EMNLP}.

\bibitem[{Zhou et~al.(2021)Zhou, Huang, Ma, and Huang}]{zhou2021document}
Wenxuan Zhou, Kevin Huang, Tengyu Ma, and Jing Huang. 2021.
\newblock \href {https://ojs.aaai.org/index.php/AAAI/article/view/17717}
  {Document-level relation extraction with adaptive thresholding and localized
  context pooling}.
\newblock In \emph{Proceedings of AAAI}.

\end{thebibliography}

\end{document}